\documentclass[10pt,twocolumn,letterpaper]{article}

\pdfoutput=1

\usepackage{icb}
\usepackage{times}
\usepackage{epsfig}
\usepackage{graphicx}
\usepackage{amsmath}
\usepackage{amssymb}
\usepackage{subcaption}
\usepackage{multirow}
\usepackage{tabularx}
\usepackage{booktabs}

\icbfinalcopy % *** Uncomment this line for the final submission

\ificbfinal\fi

\begin{document}

%%%%%%%%% TITLE
\title{Video Face Recognition: Component-wise Feature Aggregation Network (C-FAN)}

\author{
    Sixue Gong\quad\quad
    Yichu Shi\quad\quad
    Anil K. Jain\quad\quad\\
    Michigan State University, East Lansing, MI\\
{\tt\small \{gongsixu,shiyichu\}@msu.edu, jain@cse.msu.edu }
}

\maketitle
%\thispagestyle{mypage}

% \pagestyle{fancy}
% \rhead{\thepage}
% \fancyfoot{}
% \cfoot{\tiny{\text \IEEEpubid{\makebox[\columnwidth]{\hfill \footnotesize 978-1-7281-3640-0/19/\$31.00 \copyright 2019 IEEE\hfill} \hspace{\columnsep}\makebox[\columnwidth]{ }} }}

%%%%%%%%% ABSTRACT
\begin{abstract}

We propose a new approach to video face recognition. Our component-wise feature aggregation network (C-FAN) accepts a set of face images of a subject as an input, and outputs a single feature vector as the face representation of the set for the recognition task. The whole network is trained in two steps: (i) train a base CNN for still image face recognition; (ii) add an aggregation module to the base network to learn the quality value for each feature component, which adaptively aggregates deep feature vectors into a single vector to represent the face in a video. C-FAN automatically learns to retain salient face features with high quality scores while suppressing features with low quality scores. The experimental results on three benchmark datasets, YouTube Faces \cite{wolf2011face}, IJB-A \cite{klare2015pushing}, and IJB-S \cite{Kalka2018IJBS} show that the proposed C-FAN network is capable of generating a compact feature vector with 512 dimensions for a video sequence by efficiently aggregating feature vectors of all the video frames to achieve state of the art performance.

\end{abstract}

% \let\thefootnote\relax\footnotetext{\mycopyrightnotice}

%%%%%%%%% BODY TEXT

\section{Introduction}

Video-based face recognition has received increasing interest in recent years as a growing number of videos are continually being captured by mobile devices and CCTV systems for surveillance. Due to requirements in law enforcement and the fact that videos contain rich temporal and multi-view information, it is necessary to develop robust and accurate face recognition techniques for surveillance systems. Although the ubiquity of deep learning algorithms has advanced face recognition technology for static face images, video-based face recognition still poses a significant research challenge. Compared to static photos which are generally taken under controlled conditions (in terms of pose, illumination, expression, and occlusion) and with subject's cooperation, individual video frames have relatively low image quality because of unconstrained capture 

environments \cite{Kalka2018IJBS}.

\begin{figure}
    \captionsetup{font=footnotesize}
    \centering
    \begin{subfigure}[b]{0.33\linewidth}
    \includegraphics[width=\linewidth]{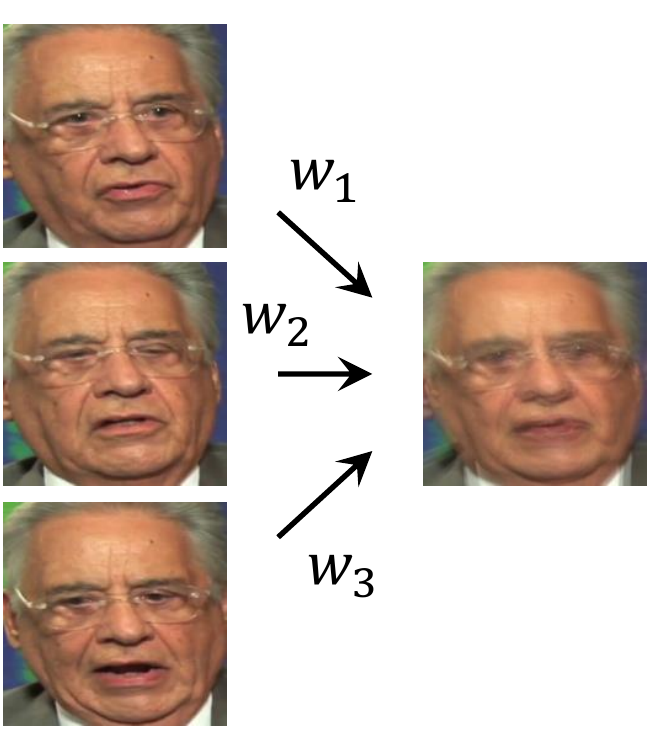}
    \caption{{\footnotesize Image-level}}
    \label{fig:image_fusion}
    \end{subfigure}\hfill
    \begin{subfigure}[b]{0.33\linewidth}
    \includegraphics[width=\linewidth]{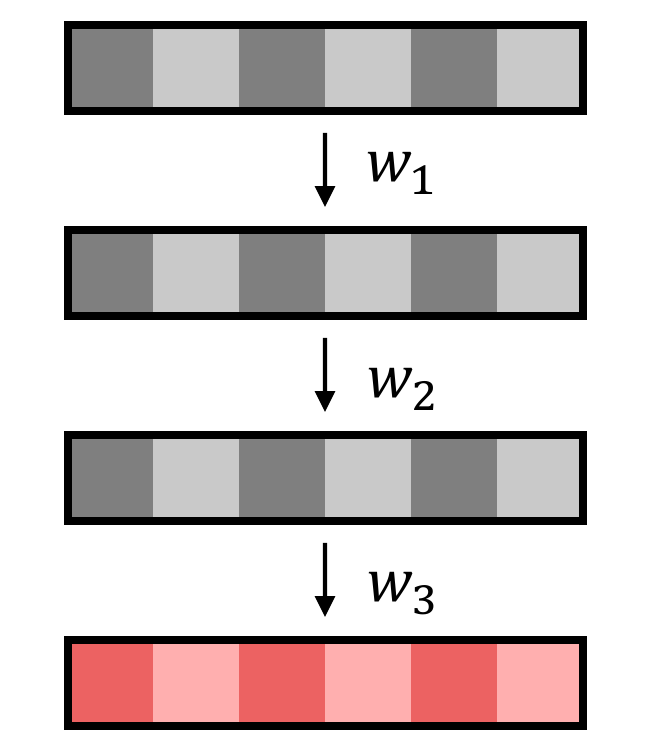}
    \caption{{\footnotesize Representation-level}}
    \label{fig:feature_vector_fusion}
    \end{subfigure}\hfill
    \begin{subfigure}[b]{0.33\linewidth}
    \includegraphics[width=\linewidth]{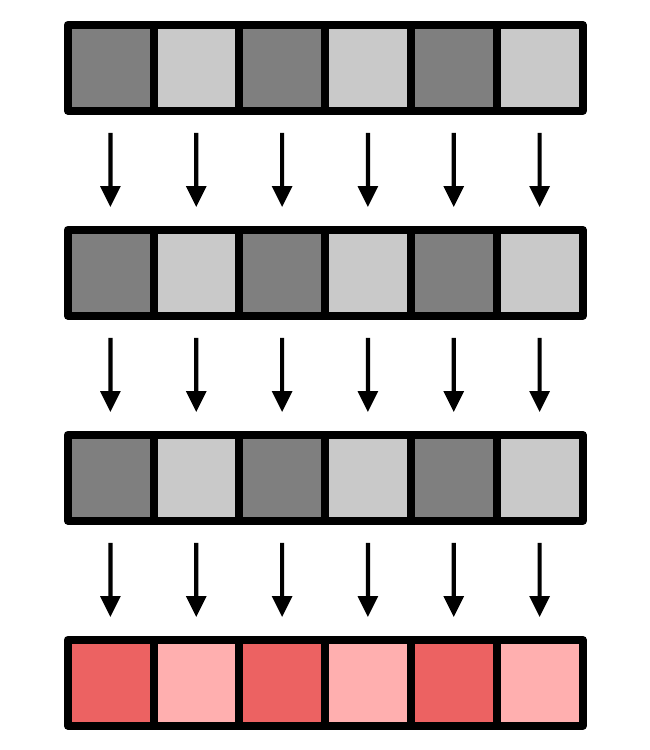}
    \caption{{\footnotesize Component-wise}}
    \label{fig:dimension_fusion}
    \end{subfigure}\\
    \caption{Methods to aggregate identity information from a set of video frames of the same person. Figure \ref{fig:image_fusion}: individual frames are merged (e.g., pixel wise pooling or generative model to obtain a fused image); Figure \ref{fig:feature_vector_fusion}: deep feature vector for each video frame is assigned a single weight, $W_i$; Figure \ref{fig:dimension_fusion}: weighted average of feature components, where weights (quality value) are learned by the same network which gives the feature vector (see Figure \ref{fig: framework})}
    \label{fig:fusion}
\end{figure}

\begin{figure*}
\captionsetup{font=footnotesize}
    \centering
    \newcommand{\vshrink}{\vspace{-10px}}
    \begin{subfigure}[b]{0.24\linewidth}
    \includegraphics[width=0.25\linewidth]{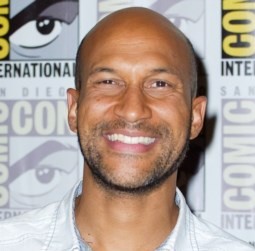}\hfill
    \includegraphics[width=0.25\linewidth]{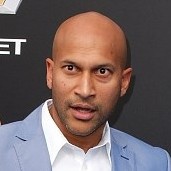}\hfill
    \includegraphics[width=0.25\linewidth]{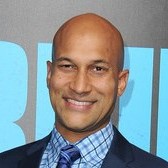}\hfill
    \includegraphics[width=0.25\linewidth]{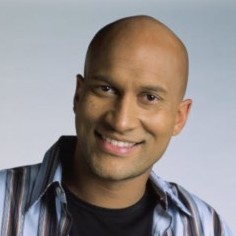}\\
    \includegraphics[width=0.25\linewidth]{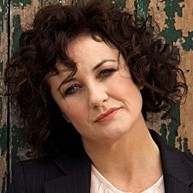}\hfill
    \includegraphics[width=0.25\linewidth]{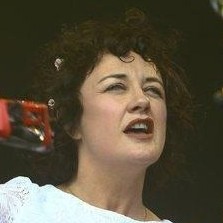}\hfill
    \includegraphics[width=0.25\linewidth]{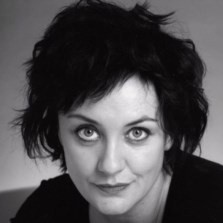}\hfill
    \includegraphics[width=0.25\linewidth]{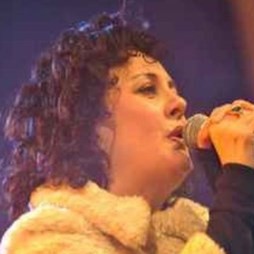}\\
    \includegraphics[width=0.25\linewidth]{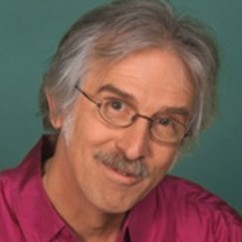}\hfill
    \includegraphics[width=0.25\linewidth]{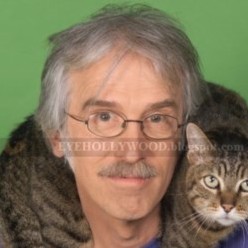}\hfill
    \includegraphics[width=0.25\linewidth]{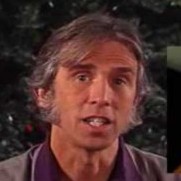}\hfill
    \includegraphics[width=0.25\linewidth]{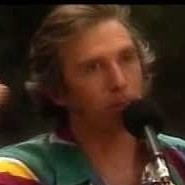}\\
    \vshrink\caption{MS-Celeb-1M \cite{guo2016msceleb}}
    \label{fig:img_msceleb}
    \end{subfigure}\hfill
    \begin{subfigure}[b]{0.24\linewidth}
    \includegraphics[width=0.25\linewidth]{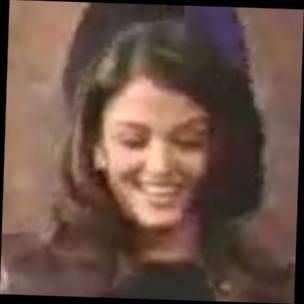}\hfill
    \includegraphics[width=0.25\linewidth]{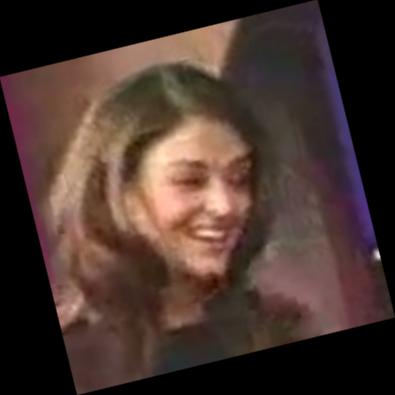}\hfill
    \includegraphics[width=0.25\linewidth]{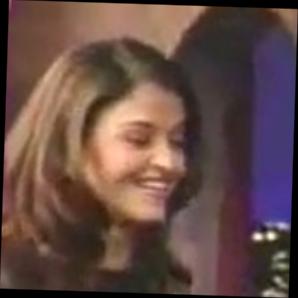}\hfill
    \includegraphics[width=0.25\linewidth]{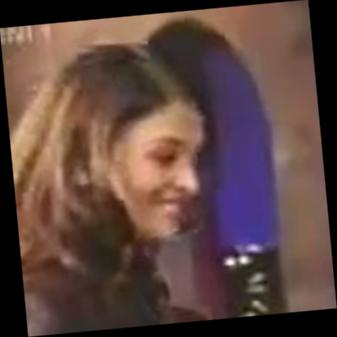}\\
    \includegraphics[width=0.25\linewidth]{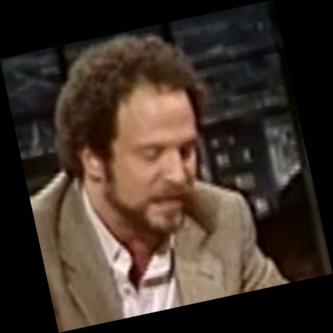}\hfill
    \includegraphics[width=0.25\linewidth]{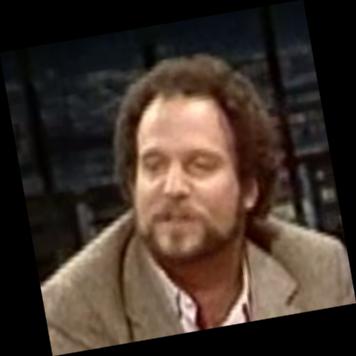}\hfill
    \includegraphics[width=0.25\linewidth]{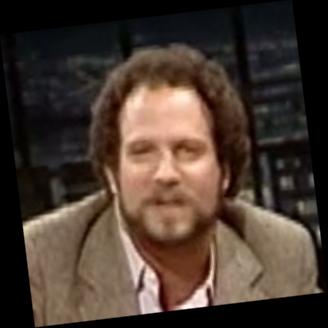}\hfill
    \includegraphics[width=0.25\linewidth]{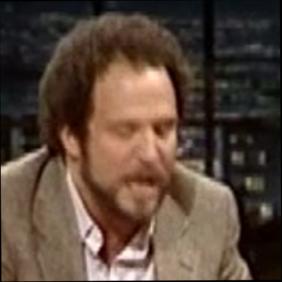}\\
    \includegraphics[width=0.25\linewidth]{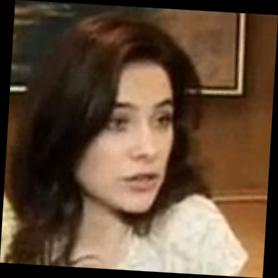}\hfill
    \includegraphics[width=0.25\linewidth]{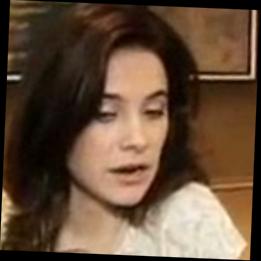}\hfill
    \includegraphics[width=0.25\linewidth]{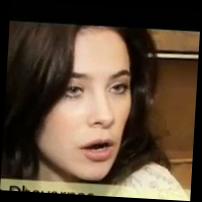}\hfill
    \includegraphics[width=0.25\linewidth]{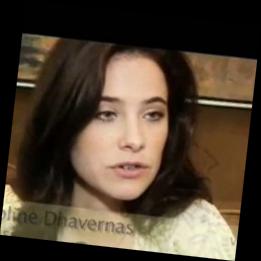}\\
    \vshrink\caption{YoutubeFaces \cite{wolf2011face}}
    \end{subfigure}\hfill
    \begin{subfigure}[b]{0.24\linewidth}
    \includegraphics[width=0.25\linewidth]{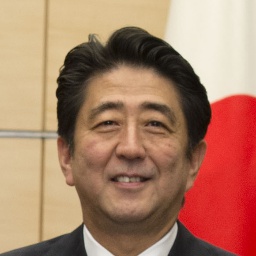}\hfill
    \includegraphics[width=0.25\linewidth]{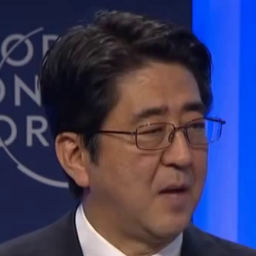}\hfill
    \includegraphics[width=0.25\linewidth]{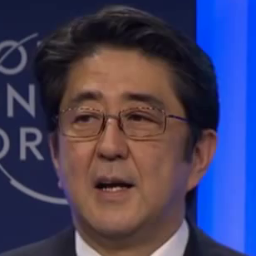}\hfill
    \includegraphics[width=0.25\linewidth]{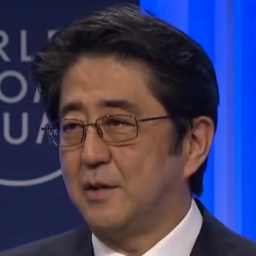}\\
    \includegraphics[width=0.25\linewidth]{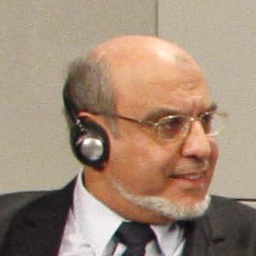}\hfill
    \includegraphics[width=0.25\linewidth]{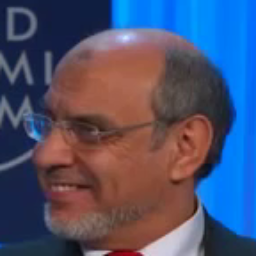}\hfill
    \includegraphics[width=0.25\linewidth]{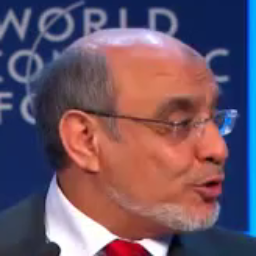}\hfill
    \includegraphics[width=0.25\linewidth]{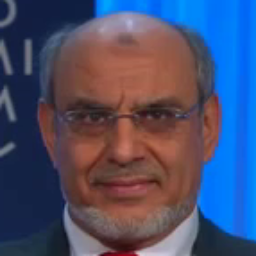}\\
    \includegraphics[width=0.25\linewidth]{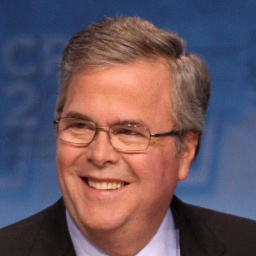}\hfill
    \includegraphics[width=0.25\linewidth]{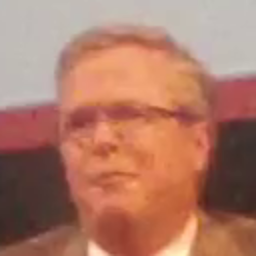}\hfill
    \includegraphics[width=0.25\linewidth]{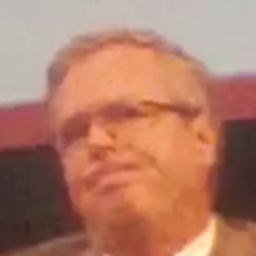}\hfill
    \includegraphics[width=0.25\linewidth]{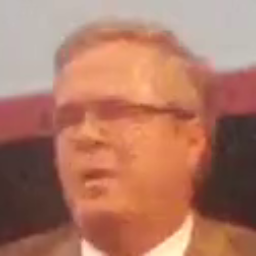}\\
    \vshrink\caption{IJB-A \cite{klare2015pushing}}
    \end{subfigure}\hfill
    \begin{subfigure}[b]{0.24\linewidth}
    \includegraphics[width=0.25\linewidth]{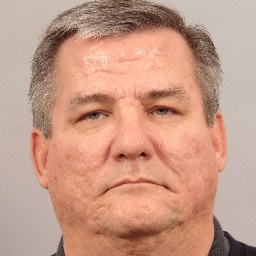}\hfill
    \includegraphics[width=0.25\linewidth]{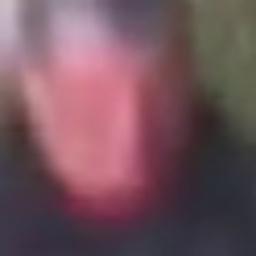}\hfill
    \includegraphics[width=0.25\linewidth]{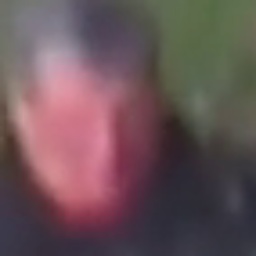}\hfill
    \includegraphics[width=0.25\linewidth]{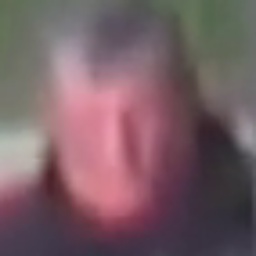}\\
    \includegraphics[width=0.25\linewidth]{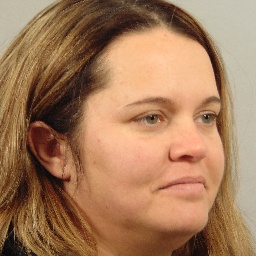}\hfill
    \includegraphics[width=0.25\linewidth]{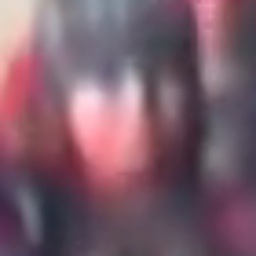}\hfill
    \includegraphics[width=0.25\linewidth]{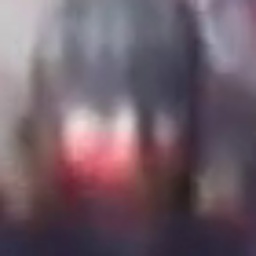}\hfill
    \includegraphics[width=0.25\linewidth]{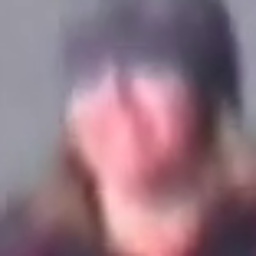}\\
    \includegraphics[width=0.25\linewidth]{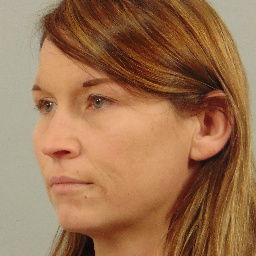}\hfill
    \includegraphics[width=0.25\linewidth]{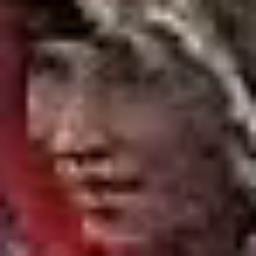}\hfill
    \includegraphics[width=0.25\linewidth]{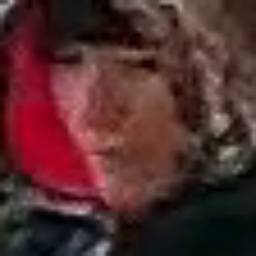}\hfill
    \includegraphics[width=0.25\linewidth]{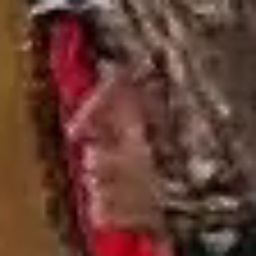}\\
    \vshrink\caption{IJB-S \cite{Kalka2018IJBS}}
    \label{fig:img_ijbs}
    \end{subfigure}\\
    \caption{Example images from different datasets. MS-Celeb-1M and YoutubeFaces contain only still images or video frames, respectively. IJB-A and IJB-S include both still images and videos. The first columns of IJB-A and IJB-S show still images, followed by video frames of the respective subjects in the next three columns.}
    \label{fig:dataset}
\end{figure*}

There are two main challenges in face recognition in surveillance videos. First, subjects captured by surveillance cameras are usually moving without directly looking at the cameras, leading to large pose variations and severe motion blur. Second, the video consists of multiple frames of the same subject, providing both noisy frames with poor quality and unfavorable viewing angles, but useful temporal and multi-view information (see Figure \ref{fig:img_ijbs}). Therefore, efficiently using a set of frames to generate a compact face representation is another challenge in video-based face recognition.

State-of-the-art face recognition algorithms for still images strive to generate robust face representations by deep neural networks trained on a large face dataset \cite{Schroff_2015_CVPR}, \cite{taigman2014deepface}, \cite{liu2017sphereface}, \cite{sun2015deeply}. Although the problems of image blur, pose variations, and occlusions can be partially solved by data augmentation \cite{ding2018trunk} and ensemble convolutional neural networks \cite{ding2018trunk}, \cite{sun2015deeply}, \cite{liu2015targeting}, such strategies may not be directly applicable to video frames. Furthermore, the majority of  existing deep face recognition approaches cast network predictions as point estimates, which means every face image has an explicit representation regardless of quality. However, there are many noisy video frames with low facial information content. For those frames, the corresponding face representation should have lower reliability. Also, in many practical surveillance applications, video clips of a range of frames for each identity can be collected. It is crucial to effectively integrate the information across different frames together from frame-level features \cite{liu2017quality}, \cite{yang2017neural} or raw input frames \cite{hassner2016pooling}, \cite{rao2017learning} by assigning each instance\footnote{An image or a feature vector.} with an adaptive instance-level weight. However, since each component of an instance encodes different subsets of facial features, noise could also be integrated when we emphasize or suppress all components simultaneously.

In this work, we propose a new approach for video face recognition by considering component-wise feature aggregation. We cast video face recognition as a template (a set of images from the same person) matching problem \footnote{The same setting as the IARPA Janus Benchmark~\cite{klare2015pushing}}. Figure \ref{fig:fusion} shows the proposed component-wise feature aggregation, compared to the other two instance-level aggregation methods. 
% For aggregation at instance-level, each instance, either an image or a feature vector, has a weight scalar. And the aggregated instance is the linear combination between the instances and the corresponding weights. Figure \ref{fig:feature_vector_fusion} and Figure \ref{fig:image_fusion} present the aggregation of feature vectors and input images respectively.
In component-wise aggregation, each component of the feature vectors is aggregated separately by learning the corresponding quality weight vectors as shown in Figure \ref{fig:dimension_fusion}.
Based on the hypothesis that 
% each component of the deep face feature vectors is independent and 
different covariates only lead to a variation in a subset of deep feature components, we train the network to predict an adaptive quality score for each deep feature component. The network can be trained on a still image dataset (e.g. MS-Celeb-1M \cite{guo2016msceleb}, Figure \ref{fig:img_msceleb}) with randomly generated templates. During testing, the features of each video or template are aggregated independently on each component using the respective quality scores. Experimental results on the IJB-S dataset \cite{Kalka2018IJBS} as well as other template/video matching benchmarks show that the proposed method significantly boosts the performance compared with average pooling and outperforms instance-level feature aggregation. The contributions of the paper are summarized below:
\begin{itemize}
    \item A component-wise feature aggregation network (C-FAN) that aggregates each component of deep feature vectors separately, leading to better representation for video face recognition. \vspace{-5px}
    \item The quality scores predicted by C-FAN correlate with the visual image quality. \vspace{-5px}
    \item Evaluating the proposed C-FAN on a challenging surveillance dataset IJB-S~\cite{Kalka2018IJBS}. Consequently, we achieve state-of-the-art results on the five face identification protocols.
    \item We also attain comparable performance on the other two face recognition benchmarks, YouTube Faces \cite{wolf2011face}, and IJB-A \cite{klare2015pushing}.
\end{itemize}

%-------------------------------------------------------------------------

%------------------------------------------------------------------------
\section{Related Work}

\subsection{Deep Face Recognition}
Deep neural networks dominate the ongoing research in face recognition ~\cite{taigman2014deepface},~\cite{deepid2},~\cite{Schroff_2015_CVPR},~\cite{masi2016we},~\cite{liu2017sphereface},~\cite{hasnat2017deepvisage},~\cite{ranjan2017l2},~\cite{wang2018additive}, given their success in the ImageNet competition ~\cite{krizhevsky2012imagenet}. Taigman~\etal~\cite{taigman2014deepface} proposed the first face recognition application using deep neural networks. The following works have been exploring different loss functions to improve the discriminability of the embedding vector. Wen~\etal ~\cite{wen2016discriminative} proposed center loss to reduce the intra-class variation. Other work proposed new metric learning methods for face recognition ~\cite{Schroff_2015_CVPR}~\cite{sohn2016improved}. More recent work has been trying to close the gap between metric learning and identification learning by learning classification on a spherical embedding space~\cite{liu2017sphereface}~\cite{wang2018additive},~\cite{wang2018cosface},~\cite{ranjan2017l2}.

\subsection{Video Face Recognition}

Previous work on video face recognition can mainly be categorized into three categories: \emph{space}-based, \emph{classifier}-based method and \emph{aggregation}-based. \emph{Space}-based methods aim to model the feature space spanned by the instances in the video, such as probabilistic distributions~\cite{shakhnarovich2002face}, ~\cite{arandjelovic2005face}, affine hulls~\cite{shakhnarovich2002face},~\cite{yang2013face},~\cite{cevikalp2010face}, SPD matrices~\cite{huang2015log}, $n$-order statistics~\cite{lu2013image} and manifolds~\cite{lee2003video},~\cite{harandi2011graph}.  \emph{Classifier}-based methods train a supervised classifier on each image set or video to obtain a representation~\cite{wolf2011face},~\cite{crosswhite2017template}.
Because deep convolution neural networks are shown to be able to lead to highly discriminative compact face representations from a collection of images, recent works on video face recognition mainly use an aggregation-based method to fuse a set of feature vectors or images into a single vector. Compared to previous methods as well as score-level fusion~\cite{best2014unconstrained}, aggregation methods are more efficient in terms of both computation and storage. In particular, image-aggregation methods fuse a set of images to a single image~\cite{hassner2016pooling},~\cite{rao2017learning} for feature extraction while feature-aggregation methods fuses the features extracted from different images into one vector. Although some feature-aggregation methods achieve good performance by simply using an average pooling~\cite{ding2018trunk},~\cite{chen2018unconstrained}, a number of recent work has focused on fusing features with visual qualities, such as detection score~\cite{ranjan2018crystal}, or predicted quality scores~\cite{liu2017quality},~\cite{yang2017neural},~\cite{sankaran2018metadata}. However, all of them only consider an instance-level aggregation while our work tries to aggregate the feature vectors in each component separately.

% \emph{Instance}-based methods aim to select feature-rich frames from the videos to represent and compare the videos~\cite{goswami2017face}.
% Rao\etal~\cite{rao2017attention} employ reinforcement to learn attention-based method for learning representation from temporal data.
% Bodla~\etal~\cite{bodla2017deep} proposed to pass averaged features to a fusion network to learn a new feature vector that is concatenated with the original one.

\section{Component-wise Feature Aggregation}

% In this paper, we proposed a Dimension-wise Feature Aggregation Network (C-FAN) whose input is a set of face images of a subject, i.e., a template, and outputs a single feature vector as the face representation for the recognition task. It has two modules, feature embedding module and feature aggregation module, that are implemented through an end-to-end deep convolutional neural network as shown in Figure \ref{fig: framework}. The proposed C-FAN network is capable of generating a compact feature vector of 512-dimension for a template by adaptively aggregating features of all frames in a video while maintaining the discriminative power at the same time.

% In this section, we first conduct a few experiments to demonstrate the motivation of the proposed component-wise feature aggregation network (C-FAN). Then we present more details of the aggregation module and its training strategy.

\subsection{Motivation}

% \begin{figure}
%     \centering
%     \newcommand{\vshrink}{\vspace{-18px}}
%     \begin{subfigure}[b]{0.5\linewidth}
%     \includegraphics[width=\linewidth]{figs/dev_downsample2x.pdf}
%     \vshrink\caption{downsampling (2x)}
%     \end{subfigure}\hfill
%     \begin{subfigure}[b]{0.5\linewidth}
%     \includegraphics[width=\linewidth]{figs/dev_motionblur15.pdf}
%     \vshrink\caption{motion blur}
%     \end{subfigure}\\
%     \begin{subfigure}[b]{0.5\linewidth}
%     \includegraphics[width=\linewidth]{figs/dev_randomnoise0_3.pdf}
%     \vshrink\caption{random noise}
%     \end{subfigure}\hfill
%     \begin{subfigure}[b]{0.5\linewidth}
%     \includegraphics[width=\linewidth]{figs/dev_rotate30.pdf}
%     \vshrink\caption{rotation}
%     \end{subfigure}\\
%     \begin{subfigure}[b]{0.5\linewidth}
%     \includegraphics[width=\linewidth]{figs/dev_translatey30.pdf}
%     \vshrink\caption{translation}
%     \end{subfigure}\hfill
%     \begin{subfigure}[b]{0.5\linewidth}
%     \includegraphics[width=\linewidth]{figs/dev_occlusion15-45.pdf}
%     \vshrink\caption{occlusion}
%     \end{subfigure}\\
%     \caption{Illustration of the deviation in individual feature component value by adding image degradation to the original high-quality images}
%     \label{fig:noise_bar}
% \end{figure}

Having multiple frames of a face in a video clip can be both advantageous and challenging at the same time. On the one hand, multiple frames incorporate temporal and multi-view information; on the other hand, large pose variations and motion blur of the frames will incur data noise and may impair the accuracy of video-based face recognition. Therefore, a key issue is how to aggregate the rich information in video frames to generate a face representation with stronger discriminative power than of individual frames.

\begin{figure}
\captionsetup{font=footnotesize}
    \centering
    \includegraphics[width=\linewidth]{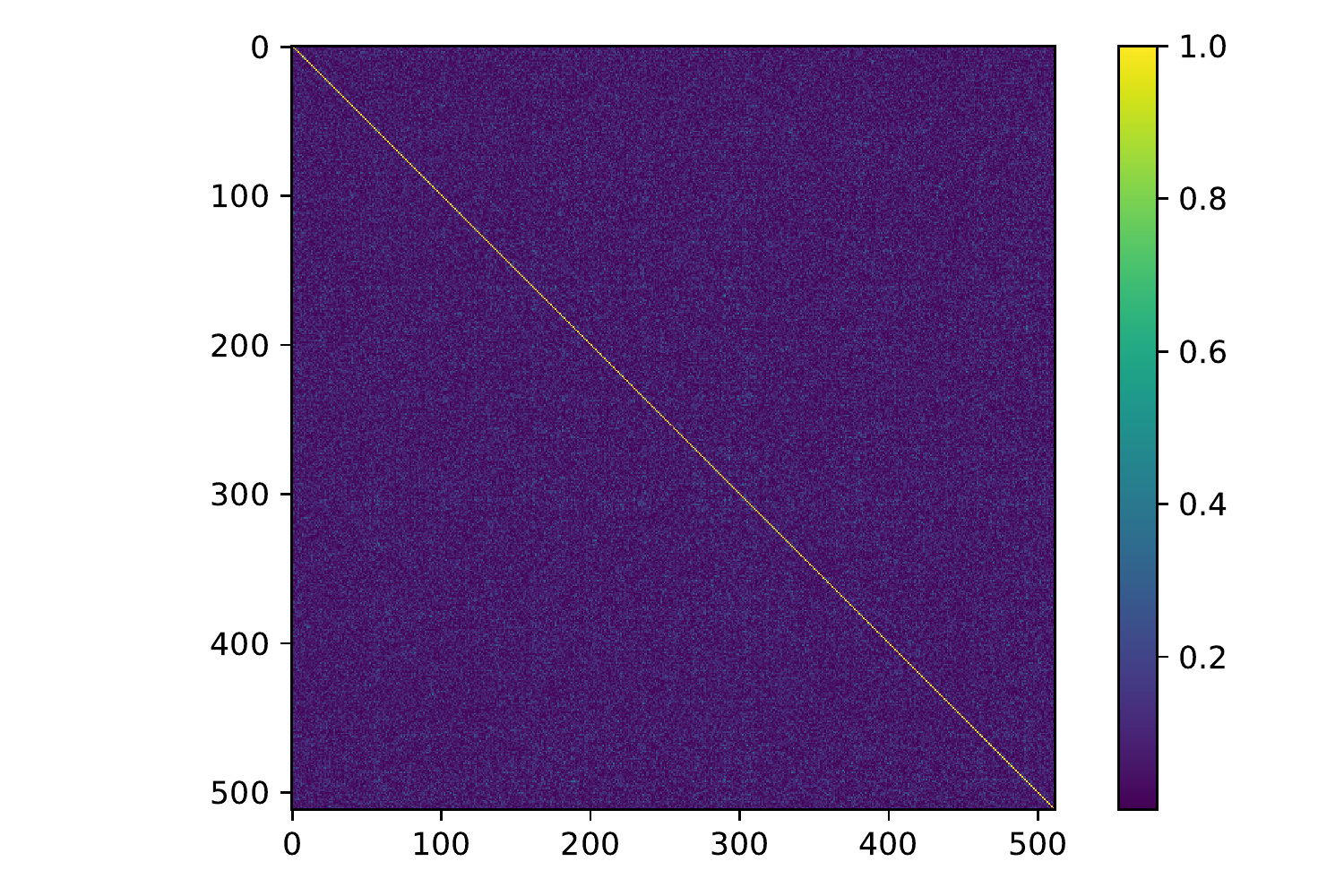}
    \caption{Heat map of the intra-class deep features extracted from the feature embedding module of C-FAN on IJB-S dataset~\cite{Kalka2018IJBS}. Instead of original correlation values, we show their absolute values because we are only interested in their magnitude.  }
    \label{fig:heatmap}
\end{figure}

Hassner~\etal~\cite{hassner2016pooling} found that simply eliminating poor quality frames will not necessarily help improve the face recognition accuracy.
One of the possible reasons is that even poor quality frames carry some valuable information. Efforts have been made to aggregate multiple frames into one single image \cite{hassner2016pooling}, \cite{rao2017learning}. A more straightforward solution might be extracting a feature vector for each frame and averaging the feature vectors to obtain a single video representation, which is known as average pooling. Let $T = \{I_1, I_2, \cdots, I_N\}$ be a template of face images. Assume that each subject has a noiseless embedding $\boldsymbol{\mu}$, and a feature vector of the $i^{th}$ image in the template $T$ generated by face representation model is
\begin{equation}
    \mathbf{f}_i = \boldsymbol{\mu} + \boldsymbol{\epsilon}_i,
\end{equation}
where $\boldsymbol{\epsilon}_i$ is the noise caused by the variations in the image. The template representation of average pooling can thus be explained as
\begin{equation}
    \mathbf{r}^{avg} = \frac{1}{N} \sum_{i}(\boldsymbol{\mu} + \boldsymbol{\epsilon}_i) = \boldsymbol{\mu} + \frac{1}{N} \sum_{i}{\boldsymbol{\epsilon}}_i,
\end{equation}
where it assumes the expectation of noise $E(\boldsymbol{\epsilon}) = \mathbf{0}$. If the number of images $N$ is big enough, the approximation of expectation $\hat{E}(\boldsymbol{\epsilon}) = \frac{1}{N} \sum_{i}{\boldsymbol{\epsilon}}_i$ will converge to $\mathbf{0}$. However, the assumption cannot hold when $N$ is small. Previous work \cite{liu2017quality}, \cite{yang2017neural} attempts to solve this problem by learning a weight scalar for each feature vector (referred to as instance) and their template representation can formulated as:
\begin{equation}
    \mathbf{r}^{inst} = \frac{1}{N} \sum_{i} w_i (\boldsymbol{\mu} + \boldsymbol{\epsilon}_i) = \boldsymbol{\mu} + \frac{1}{N} \sum_{i} w_i \boldsymbol{\epsilon}_i,
\end{equation}
where $w_i$ is the weight for the feature vector of the $i^{th}$ image. Although it can reduce noise to some extent, it still assumes that individual components of a feature vector have correlated noise such that $\frac{1}{N} \sum_{i} w_i \epsilon_{ij} < \frac{1}{N} \sum_{i}{\epsilon_{ij}}$ for all $j$, where $j$ is the index of the $j^{th}$ feature component. However, such assumption does not hold if each component contains different amount of identity information and their noise values are weakly correlated to each other. In Figure~\ref{fig:heatmap} we can observe that the intra-class correlations between individual components (off-diagonal values) of the deep feature vectors are quite small (close to $0$). This inspires us to learn component-wise feature aggregation to further minimize noise on each component. 
% Unlike the fusion of feature vectors where all components of the same vector share a single weight, our component-wise 
The proposed fusion will generate different weights for each component of the deep feature vectors, and each component of the template representation is
\begin{equation}
    r_{j}^{C-FAN} = \mu_j + \frac{1}{N} \sum_{i} w_{ij} \epsilon_{ij}.
\end{equation}

The proposed method will make better use of the components with more identity information instead of enhancing or weakening the whole feature vector if different components of a feature vector contain different identity information. 
% To provide further support to this hypothesis, we show the heat map corresponding to the $512 \times 512$ covariance matrix using intra-class feature representations for video frames in the IJB-S benchmark face dataset. Figure \ref{fig:heatmap} shows that the correlation between individual components (off-diagonal values) of the deep feature vectors is small in comparison with their variances.

% We observe that for some frames with occlusion or large pose variations, if we crop parts of the frames and put these cropped frames together, we still can obtain a holistic face image. 

% We also present the deviation of each feature component between the original high-quality images and the corresponding images with degradation. Figure \ref{fig:noise_bar} illustrates six different kinds of image degradation. As we can see, components have various responses to the data noise. Some components are influenced substantially by image degradation while others are only slightly affected.

\subsection{Overall Framework}

\begin{figure}
\captionsetup{font=footnotesize}
    \centering
    \includegraphics[width=\linewidth]{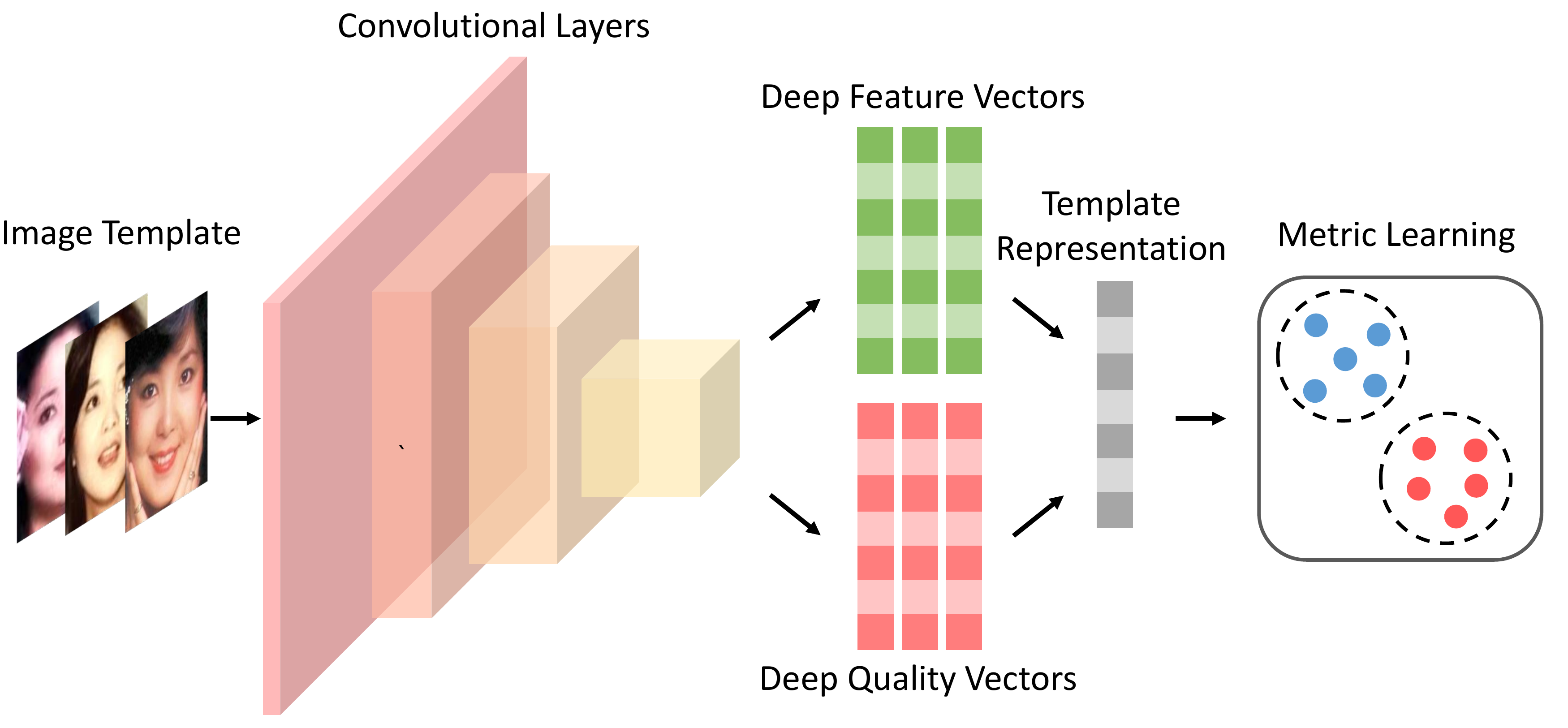}
    \caption{Framework of the proposed C-FAN.}
    \label{fig: framework}
\end{figure}

% Based on the aforementioned properties of deep feature vectors, that each component has a different response to data noise, we propose C-FAN to predict a quality score for each feature component. The components with more discriminatory information are assigned higher quality scores while the components with less information are assigned lower scores. The quality scores are then normalized to aggregate features within a template.

The overall framework of the proposed C-FAN is presented in Figure \ref{fig: framework}. C-FAN incorporates a base CNN model for extracting face representation and a feature aggregation module. The base model is first trained on a large scale face dataset (MS-Celeb-1M~\cite{guo2016msceleb}). Then we fix the base model and train the aggregation module to learn the quality score for each single feature component on the same dataset using a different objective function. The aggregation module is attached to the last convolutional layer of the base model as a second branch besides the feature extraction layer. The features of a template are then pooled with the component-wise quality scores into a compact feature vector.

% \subsection{Base CNN model}

\subsection{Feature Aggregation Module}
% Our work focuses on aggregation of a set of feature vectors from a video clip into a single vector while either maintaining or increasing the discriminative power for effective and efficient video-based face recognition. 
% Let a face template be represented by a set of images of a subject. Thus, 
% A template of the $i^{\text{th}}$ subject can be denoted as: $T^i = \{I_1^i, I_2^i, \cdots, I_N^i\}$, where $I^i_j$ is the $j^{\text{th}}$ face image of the $i^{\text{th}}$ subject, and $N$ is the total number of frames in the clip. Notice that N here is not a fixed number but changes with template size. For simplicity, we ignore the upper index $i$. All notations and discussions are within a single template. 
% Thus, each frame $I_j$ is fed into the embedding network, which produces a corresponding feature vector with fixed dimension $D$. 
Let $T = \{I_1, I_2, \cdots, I_N\}$ be a template of face images. Notice that $N$ here is not a fixed number but changes with template size. Let $\mathcal{H}(\cdot)$ denote deep convolutional layers before the feature extraction layer, and $\mathcal{F}(\cdot)$ denote the feature extraction layer. 
The feature map of $i^\text{th}$ image in the template is generated by: $\mathcal{H}(I_i) = \mathbf{m}_i$, whose feature vector is then extracted by: $\mathcal{F}(\mathbf{m}_i) = \mathbf{f}_i$, where $\mathbf{f}_i$ is a $D$-dimensional feature vector. The corresponding quality vector is obtained by $\mathcal{Q}(\mathbf{m}_i) = \mathbf{q}_i$, where $\mathcal{Q}(\cdot)$ is the feature aggregation module, and $\mathbf{q}_i$ has the same dimension $D$ as the feature vector. The quality vectors are then normalized with a softmax operator.
% to generate both positive and normalized aggregation weights $\{\mathbf{w}_1, \mathbf{w}_2, \ldots, \mathbf{w}_N\}$ with $\sum_{i=1}^{N}{w_{ij}} = 1$, where $j = \{1,2, \ldots, D\}$ is the index of the feature component. 
Formally, given a set of quality vectors $\{\mathbf{q}_1, \mathbf{q}_2, \ldots, \mathbf{q}_N\}$, the $j^{\text{th}}$ component of the $i^{\text{th}}$ vector is normalized by:
\begin{equation}
    w_{ij} = \frac{exp(q_{ij})}{\sum_{k=1}^{N}{exp(q_{kj})}}
\end{equation}

The final face representation of a template is obtained by pooling the feature vectors using the normalized quality vectors
\begin{equation}
\label{eq:fusion}
    \mathbf{r} = \sum_{i=1}^{N}{\mathbf{f}_i \odot \mathbf{w}_i},
\end{equation}
where $\odot$ represents the operation of element-wise multiplication. The aggregated feature vector of a template $\mathbf{r}$ has the same dimensionality as a single face image frame extracted by the embedding module. 
% Besides, this result is invariant to the order of input images (frames). If the template is a set of video frames, the temporal information will not influence outputs of the proposed C-FAN. 

% Moreover, the quality score for each dimension of every feature vector is optimized by the back propagation of the neural network using the loss function for face recognition. In this way, the proposed C-FAN is trained in an end-to-end manner that requires no image quality supervision.

\subsection{Training C-FAN}
Given a pre-trained CNN for face feature extraction, C-FAN adds an aggregation module to the original network to predict the quality for each component. Unlike previous work~\cite{yang2017neural} which takes the feature vector as input, we use the last feature map as the input of the aggregation layer since it contains more information. The feature aggregation module includes only one batch normalization layer and one fully connected layer. The batch normalization layer is added here to alleviate the difficulty of hyper-parameter tuning by normalizing the input of the aggregation module. During training, we fix the base CNN model and only update the aggregation layer. That is, the new model only learns how to aggregate the original features without changing them. To train the aggregation layer, we use a template triplet loss, where each triplet consists of three random templates. The templates are randomly constructed online for each mini-batch. The template features are obtained by Equation~(\ref{eq:fusion}). Online hard triplet mining is adopted and the loss function is given by:
\begin{equation}
    \mathcal{L}_{triplet} = \sum_{i=1}^{M}{[\alpha+d(\mathbf{r}^{a},\mathbf{r}^{+}) - d(\mathbf{r}^{a},\mathbf{r}^{-})]_+},
\end{equation}
where $M$ is the number of triplets, $\mathbf{r}^{a}$, $\mathbf{r}^{+}$ and $\mathbf{r}^{-}$ are the fused features of anchor template, positive template and negative template, respectively. $[x]_+=\max\{0,x\}$, $\alpha$ is a margin parameter and $d(\cdot)$ is the squared euclidean distance.
% Similarly, we also propose a template contrastive loss for the cases only paired training data are used:
% \begin{gather}
%     \mathcal{L}_{contrastive} = \sum_{i=1}^{M}{ y_id(\mathbf{r}_{i}^{1},\mathbf{r}_{i}^{2}) + (1-y_i)[\beta - d(\mathbf{r}_{i}^{1},\mathbf{r}_{i}^{2})]_+},
% \end{gather}the
% where $\mathbf{r}_{i}^{1}$ and $\mathbf{r}_{i}^{2}$ are the fused features of $i^{\text{th}}$ pair and $y_i$ is the label of the pair, i.e. $y_i=1$ for genuine pairs and $y_i=0$ for impostor pairs. $\beta$ is the margin parameter.

\section{Experiments}
% In this section, we first introduce the datasets for evaluating the proposed C-FAN. Then, we will present our training details, baseline methods, followed by a comparison of results of both C-FAN and baseline methods.

\begin{figure*}
\newcolumntype{Y}{>{\centering\arraybackslash}X}
\captionsetup{font=footnotesize}
\centering
\setlength{\arrayrulewidth}{1px}
\begin{tabularx}{\linewidth}{Y|Y|Y}
      \includegraphics[height=43px]{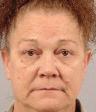}\;\includegraphics[height=43px,clip,trim={0 5px 0 5px}]{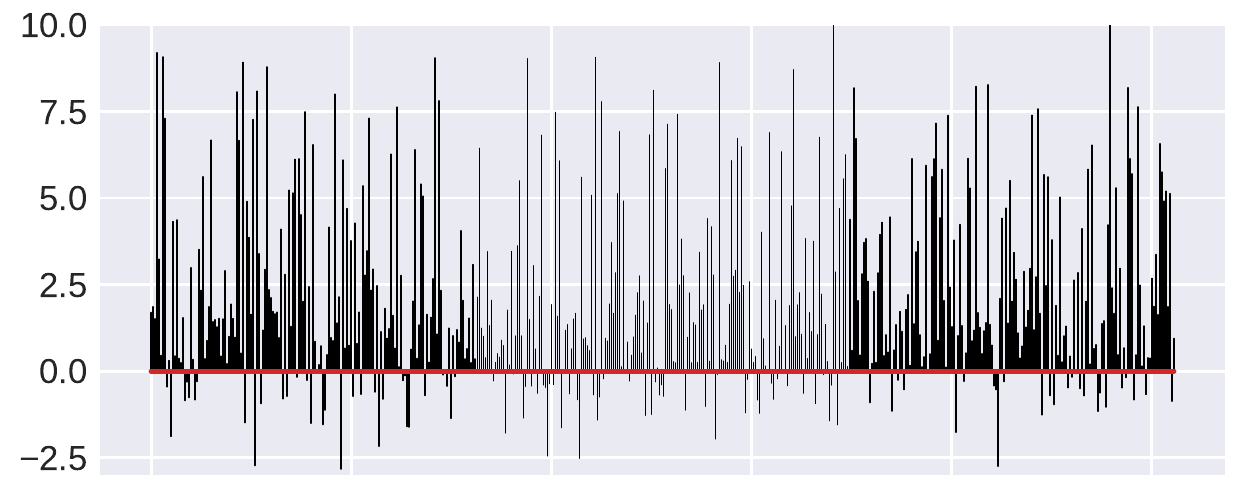}
    & \includegraphics[height=43px]{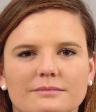}\;\includegraphics[height=43px,clip,trim={0 5px 0 5px}]{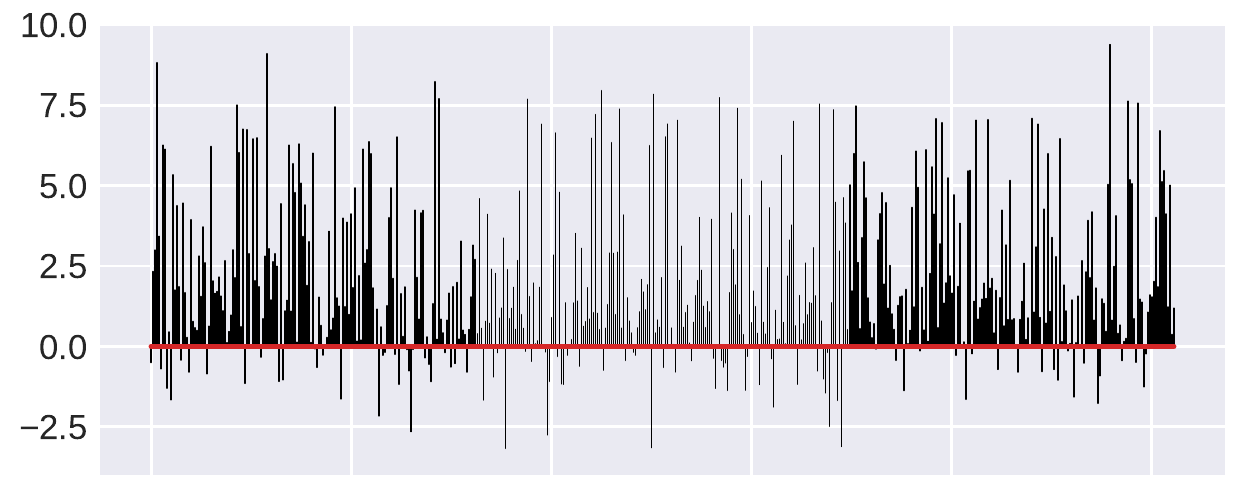} 
    & \includegraphics[height=43px]{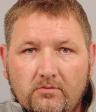}\;\includegraphics[height=43px,clip,trim={0 5px 0 5px}]{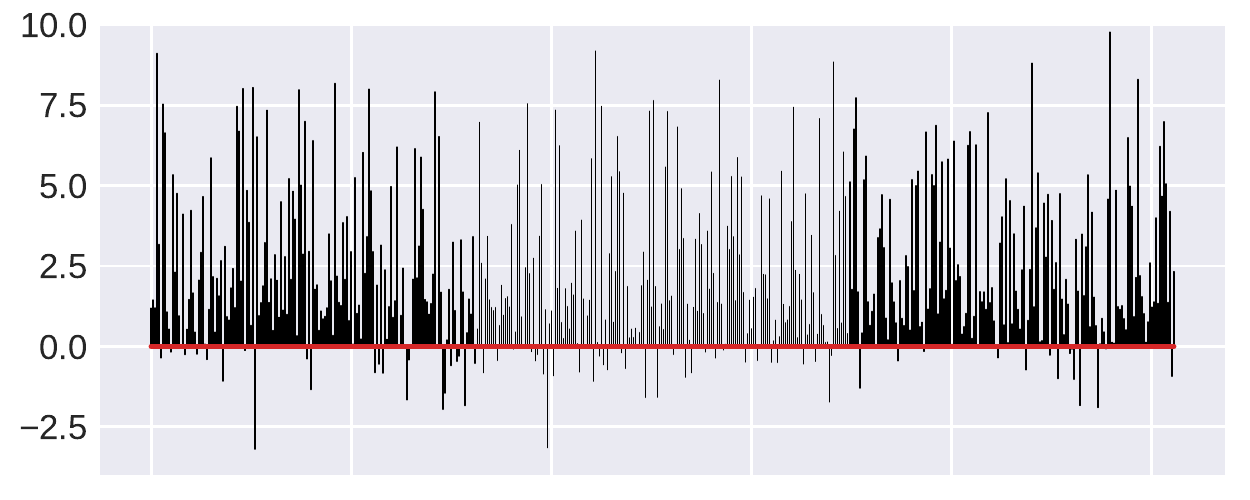}  \\
      \includegraphics[height=43px]{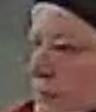}\;\includegraphics[height=43px,clip,trim={0 5px 0 5px}]{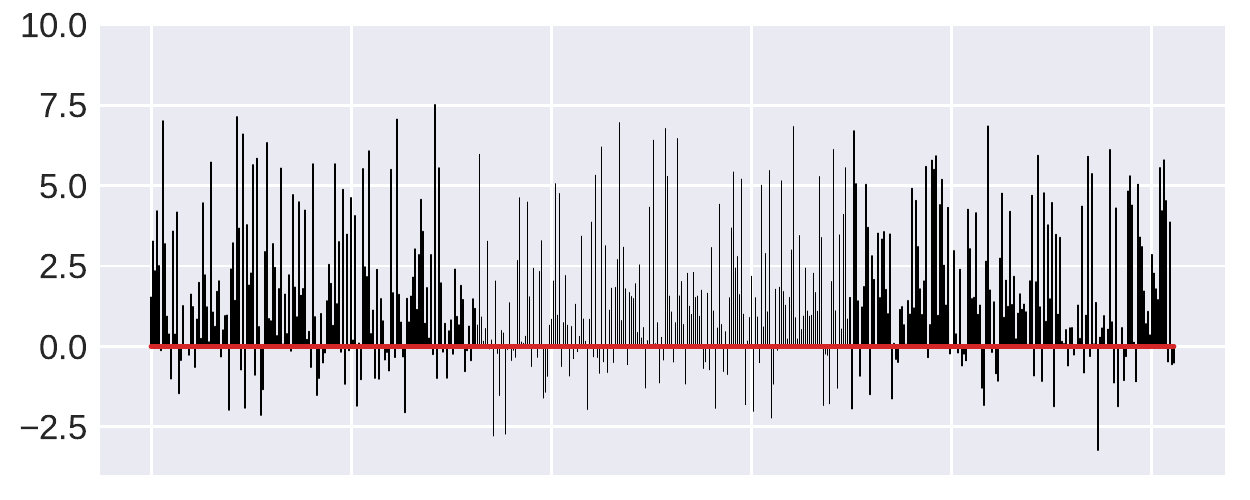}
    & \includegraphics[height=43px]{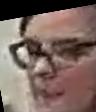}\;\includegraphics[height=43px,clip,trim={0 5px 0 5px}]{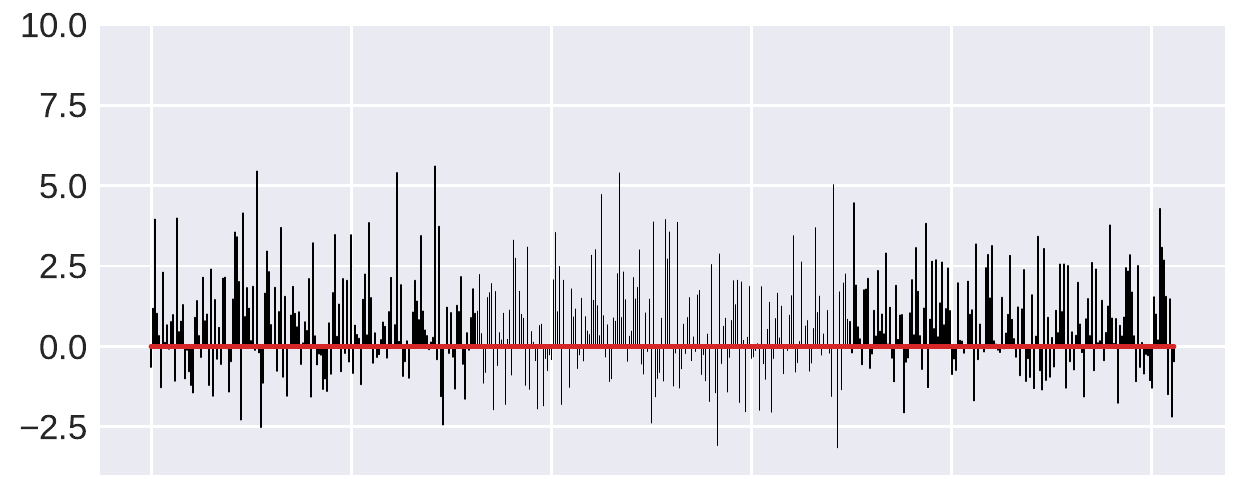} 
    & \includegraphics[height=43px]{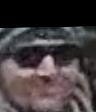}\;\includegraphics[height=43px,clip,trim={0 5px 0 5px}]{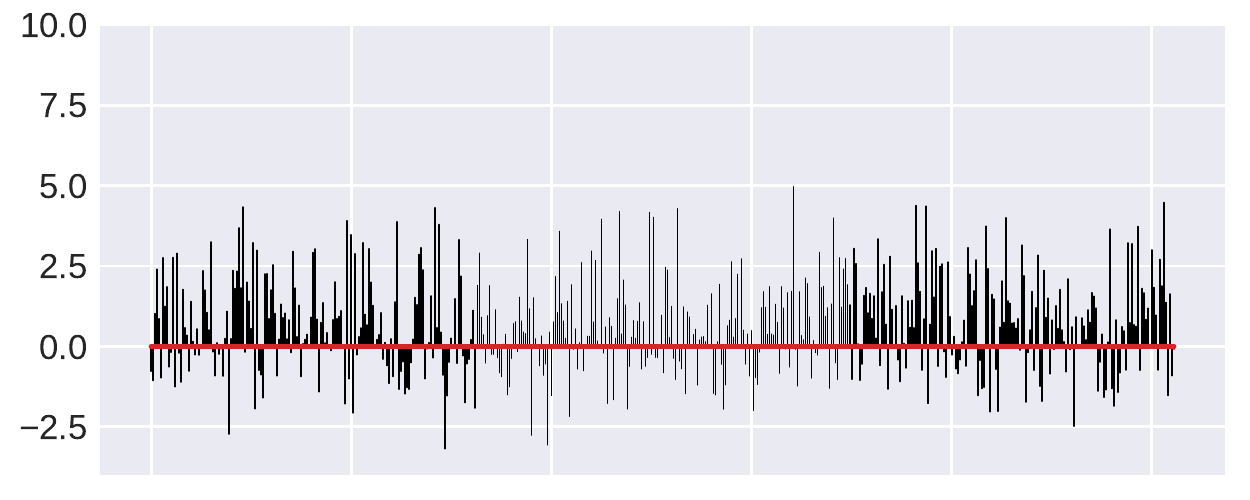}  \\
      \includegraphics[height=43px]{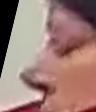}\;\includegraphics[height=43px,clip,trim={0 5px 0 5px}]{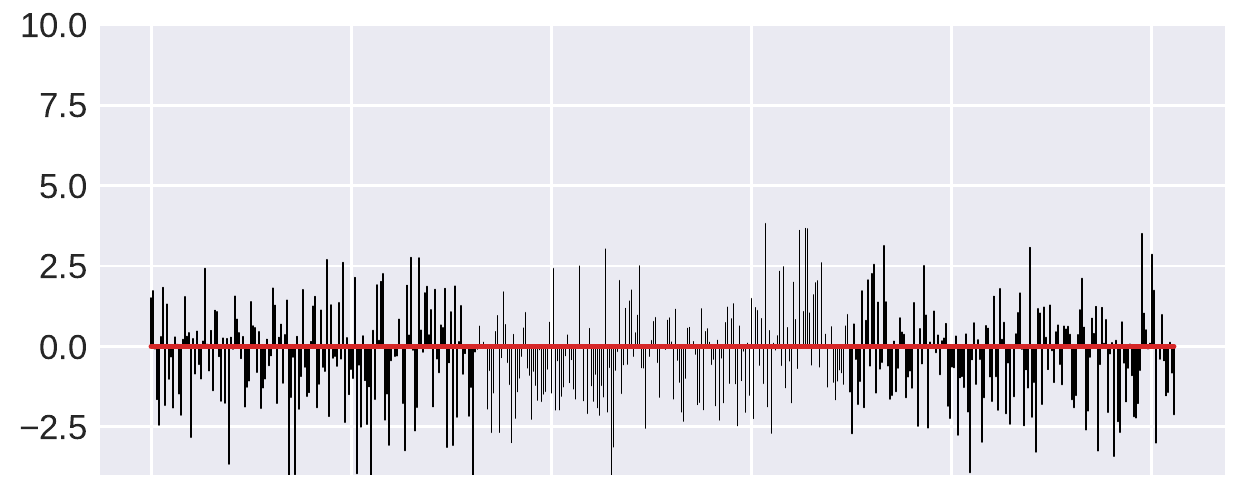} 
    & \includegraphics[height=43px]{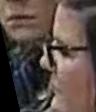}\;\includegraphics[height=43px,clip,trim={0 5px 0 5px}]{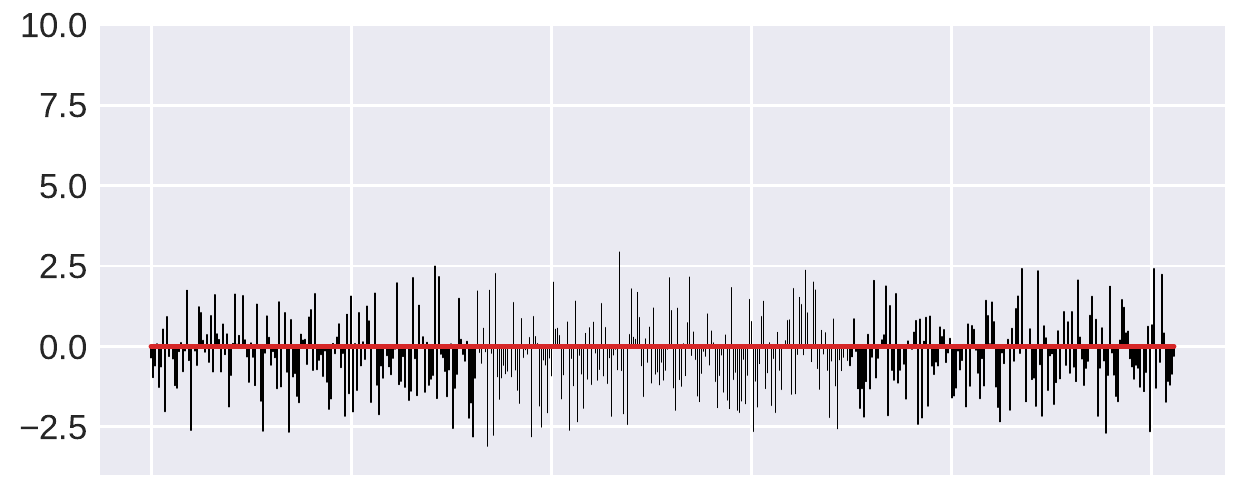} 
    & \includegraphics[height=43px]{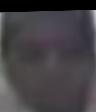}\;\includegraphics[height=43px,clip,trim={0 5px 0 5px}]{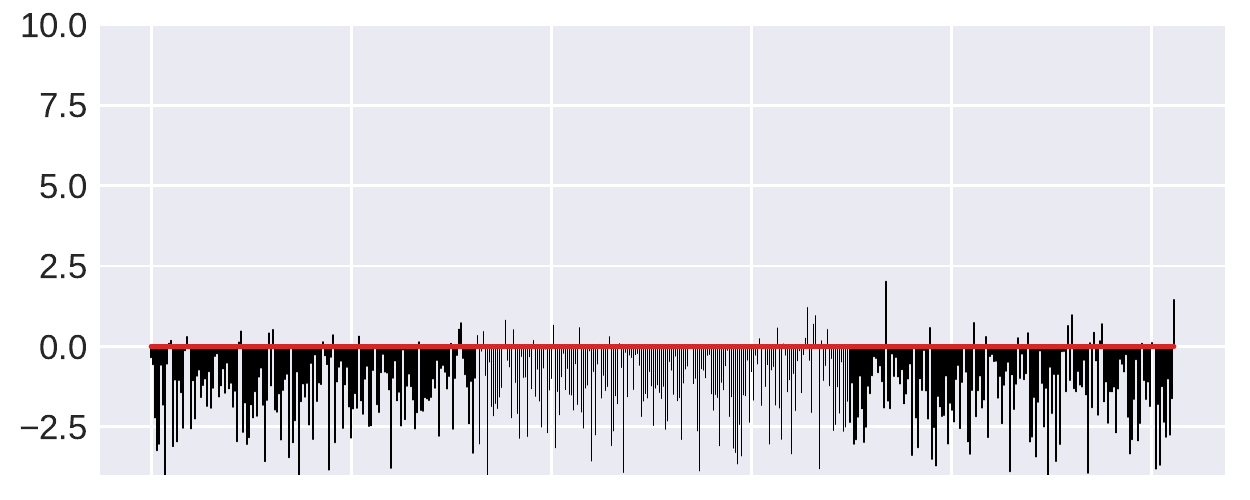}  \\
\end{tabularx}
\caption{Component-wise quality scores given by C-FAN for images of three subjects in the IJB-S dataset. Unnormalized quality scores are shown here for fair comparison. In the plots, quality values are shown for each of the 512 components. Each column shows a high-quality photo, a medium-quality frame and a low-quality frame from the same person, respectively.}
\label{fig:quality-scores}
\end{figure*}

\subsection{Datasets and Protocols}

We evaluated C-FAN on three datasets including two video face datasets, the IARPA Janus Benchmark -- Surveillance (IJB-S) \cite{Kalka2018IJBS}, the YouTube Face datast (YTF) \cite{wolf2011face}; and a template-based face dataset, the IARPA Janus Benchmark A (IJB-A) \cite{klare2015pushing}.

\textbf{\textit{IJB-S}}: The IJB-S dataset \cite{Kalka2018IJBS} is a new dataset released by IARPA for evaluating face recognition systems on surveillance videos. The dataset includes $350$ surveillance videos spanning $30$ hours in total, $5,656$ enrollment images, and $202$ enrollment videos. This dataset provides a number of challenging face videos that are captured in real-world environments with a variety of subject activities and interactions. There are five 1:N face identification evaluations (supporting both open- and closed-set evaluation) involving surveillance video, including (i) surveillance-to-still \footnote{"Still" refers to single frontal still images.}, (ii) surveillance-to-booking \footnote{The "booking" reference template comprises the full set of images captured of a single subject at enrollment time.}, (iii) Multi-view surveillance-to-booking, (iv) surveillance-to-surveillance, and (v) UAV \footnote{UAV is a small fixed-wing unmanned aerial vehicle that was flown to collect images and videos.} surveillance-to-booking. In our experiments, we follow all five identification protocols to evaluate our method. Because of the poor quality of the images in IJB-S, we are only able to detect around $9$M out of $16$M faces. In the original protocol of IJB-S, the recognition performance is normalized by the detection rate~\cite{Kalka2018IJBS}. \emph{However, since our work focuses on recognition and not detection, we do not follow the original metric. Instead, we report the standard Identification Rate (IR) and TPIR@FPIR\footnote{True positive identification and false positive identification rate.}.} Failure-to-enroll images are ignored during feature aggregation. For templates that do not contain any detected faces, we set its representation as a zero-vector.

\textbf{\textit{YTF}}: The YouTube Face dataset is a video face dataset released in 2011~\cite{wolf2011face}. It contains $3,425$ videos of $1,595$ different subjects. The number of frames in the YTF face videos ranges from $48$ to $6,070$, and the average number of frames is $181$. Compared with IJB-S, the media of YTF is more photojournalistic \cite{Kalka2018IJBS}. In experiments, we follow the 1:1 face verification protocol with the given $5,00$0 video pairs. 

\textbf{\textit{IJB-A}}: IJB-A\cite{klare2015pushing} is a template-based unconstrained face recognition benchmark. Although its images present similar challenges as IJB-S, the templates in the IJB-A include images from mixed media sources, with average image quality being better than IJB-S. The benchmark provides template-based 1:1 verification and 1:N identification protocols. IJB-A contains $500$ subjects with a total of $25,813$ images, and has been widely used by a number of both still image and video-based face recognition algorithms.

\begin{figure*}
\newcolumntype{Y}{>{\centering\arraybackslash}X}
\newcommand{\imagesize}{0.2\linewidth}
\captionsetup{font=footnotesize}
\centering
\setlength{\arrayrulewidth}{1px}
\setlength{\tabcolsep}{0.7px}
\begin{tabularx}{\linewidth}{YYY}
Probe & Retrieved & Ground-truth \\
      \includegraphics[width=\imagesize]{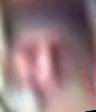}\hfill
      \includegraphics[width=\imagesize]{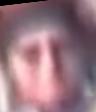}\hfill
      \includegraphics[width=\imagesize]{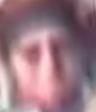}\hfill
      \includegraphics[width=\imagesize]{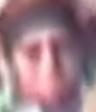}\hfill
      \includegraphics[width=\imagesize]{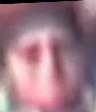}&
      \includegraphics[width=\imagesize]{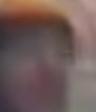}\hfill
      \includegraphics[width=\imagesize]{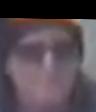}\hfill
      \includegraphics[width=\imagesize]{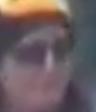}\hfill
      \includegraphics[width=\imagesize]{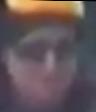}\hfill
      \includegraphics[width=\imagesize]{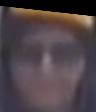}&
      \includegraphics[width=\imagesize]{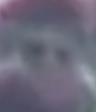}\hfill
      \includegraphics[width=\imagesize]{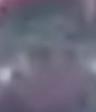}\hfill
      \includegraphics[width=\imagesize]{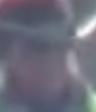}\hfill
      \includegraphics[width=\imagesize]{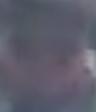}\hfill
      \includegraphics[width=\imagesize]{figs/failure_surveillance/229_196/videos_1101_129952.jpg} \\
      \includegraphics[width=\imagesize]{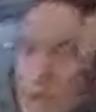}\hfill
      \includegraphics[width=\imagesize]{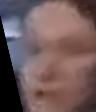}\hfill
      \includegraphics[width=\imagesize]{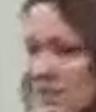}\hfill
      \includegraphics[width=\imagesize]{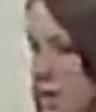}\hfill
      \includegraphics[width=\imagesize]{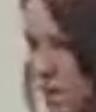} &
      \includegraphics[width=\imagesize]{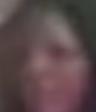}\hfill
      \includegraphics[width=\imagesize]{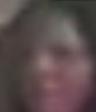}\hfill
      \includegraphics[width=\imagesize]{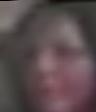}\hfill
      \includegraphics[width=\imagesize]{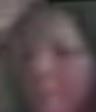}\hfill
      \includegraphics[width=\imagesize]{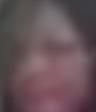}&
      \includegraphics[width=\imagesize]{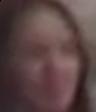}\hfill
      \includegraphics[width=\imagesize]{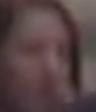}\hfill
      \includegraphics[width=\imagesize]{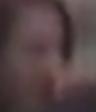}\hfill
      \includegraphics[width=\imagesize]{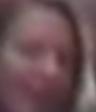}\hfill
      \includegraphics[width=\imagesize]{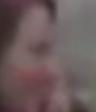} \\
      \includegraphics[width=\imagesize]{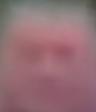}\hfill
      \includegraphics[width=\imagesize]{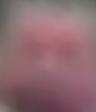}\hfill
      \includegraphics[width=\imagesize]{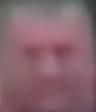}\hfill
      \includegraphics[width=\imagesize]{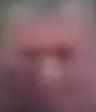}\hfill
      \includegraphics[width=\imagesize]{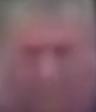}&
      \includegraphics[width=\imagesize]{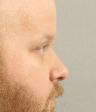}\hfill
      \includegraphics[width=\imagesize]{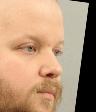}\hfill
      \includegraphics[width=\imagesize]{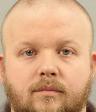}\hfill
      \includegraphics[width=\imagesize]{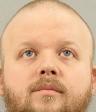}\hfill
      \includegraphics[width=\imagesize]{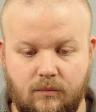}&
      \includegraphics[width=\imagesize]{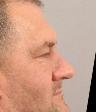}\hfill
      \includegraphics[width=\imagesize]{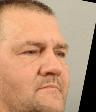}\hfill
      \includegraphics[width=\imagesize]{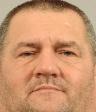}\hfill
      \includegraphics[width=\imagesize]{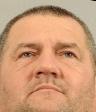}\hfill
      \includegraphics[width=\imagesize]{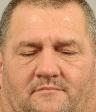}\\
      \includegraphics[width=\imagesize]{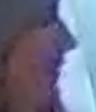}\hfill
      \includegraphics[width=\imagesize]{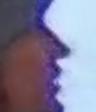}\hfill
      \includegraphics[width=\imagesize]{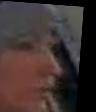}\hfill
      \includegraphics[width=\imagesize]{figs/failure_booking/183_112/videos_5081_11747.jpg}\hfill
      \includegraphics[width=\imagesize]{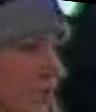}&
      \includegraphics[width=\imagesize]{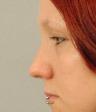}\hfill
      \includegraphics[width=\imagesize]{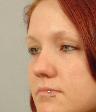}\hfill
      \includegraphics[width=\imagesize]{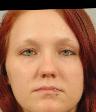}\hfill
      \includegraphics[width=\imagesize]{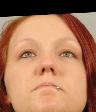}\hfill
      \includegraphics[width=\imagesize]{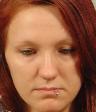}&
      \includegraphics[width=\imagesize]{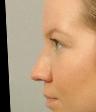}\hfill
      \includegraphics[width=\imagesize]{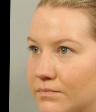}\hfill
      \includegraphics[width=\imagesize]{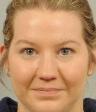}\hfill
      \includegraphics[width=\imagesize]{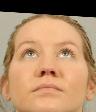}\hfill
      \includegraphics[width=\imagesize]{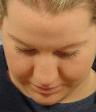}\\
\end{tabularx}
\caption{Examples of failed templates on the IJB-S dataset. The first two rows show the failure cases from "surveillance-to-surveillance" protocol. The last two show the failure cases from "surveillance-to-booking" protocol. Because of space limit, we only show 5 randomly selected images for each template.}
\label{fig:failure-examples}
\end{figure*}

\subsection{Implementation Details}
\textbf{\textit{Pre-processing}}: We employ a facial landmark detection algorithm MTCNN \cite{zhang2016joint} to detect and align all faces in training and testing. Each face is cropped from the detected face region and resized to $112 \times 96$ pixels using a similarity transformation based on the detected five facial landmarks.

% Face images are represented by a pre-trained Covolutional Neural Network (CNN) model for still image based face recognition. The CNN model is based on the Face-ResNet (DeepVisage) architecture~\cite{hasnat2017deepvisage}. We train the network on a cleaned version\footnote{\url{https://github.com/AlfredXiangWu/face_verification_experiment}.} of the MS-Celeb-1M dataset~\cite{guo2016msceleb}, containing $5$M images of $100$K subjects. The model is trained using the AM-Softmax loss function~\cite{wang2018additive}. The output of the base model is a $512$-dimensional feature vector.

\textbf{\textit{Training}}: We first train a base CNN model on a cleaned version\footnote{\url{https://github.com/AlfredXiangWu/face\_verification\_experiment}.} of MS-Celeb-1M dataset~\cite{guo2016msceleb} to learn face recognition of still images, for which we adopts the Face-ResNet (DeepVisage) architecture~\cite{hasnat2017deepvisage}. The component-wise quality module is then trained on the same training dataset using the template triplet loss. The parameters of the aggregation layer are optimized using stochastic gradient descent with a momentum of $0.9$ and a weight decay of $0.01$. Gaussian blur and motion blur are randomly applied to augment the training data. We set $\alpha=1.0$. Each mini-batch samples $240$ images of $80$ templates from $20$ random subjects, $3$ images per template. The module is trained for $4,000$ steps in all. All these hyper-parameters are chosen by testing on a validation set, which is a small subset of IJB-S. All the experiments are conducted on a Nvidia Geforce GTX 1080 Ti GPU. The average feature extraction speed on this GPU is $1$ms per image. After training on MS-Celeb-1M, the feature network and aggregation module are applied to the testing datasets without fine-tuning.

\subsection{Baseline}
We compare the proposed component-wise pooling with the other two pooling types, average pooling and instance-level pooling, in particular we compare between three methods:
\begin{itemize}
\vspace{-0.6em}
    \item \textbf{Average} directly applies the average pooling to the base CNN features to generate the template representation.\vspace{-0.6em}
    \item \textbf{Instance} is trained similarly as C-FAN, but it only outputs one quality scalar for each image and aggregates the base CNN feature at the instance-level.\vspace{-0.6em}
    \item \textbf{C-FAN} generates a quality vector for each image and does component-wise aggregation.\vspace{-0.6em}
\end{itemize}
% Since the goal of all the aggregation strategies is to generate a compact representation for a face template, we compare template-to-template similarities by leveraging pairwise distances between the aggregated feature vectors. In our experiments, 
After the aggregation, cosine similarity is used to compare the template representations.
% All the image-level feature vectors are first normalized. Next, they are aggregated through different pooling schemes to generate a template representation. And then the similarity is measured by directly computing the Euclidean distance among these template representations.
% For all the baseline methods, we use the same base CNN model of the proposed C-FAN to extract image-level feature vectors. Here, we simply use Average and Instance to refer to average pooling and instance-level aggregation, respectively. The instance here refers to image-level deep feature vectors, where the network learns a scalar quality score for each face image representation. Similar to C-FAN, both these two methods produce a 512-dimensional representation for each template.

\subsection{Qualitative Analysis on IJB-S}

% In the proposed C-FAN, we add one aggregation module to the base CNN to predict the quality value for each component of a deep feature vector. 
To explore the relationship between visual quality and the quality scores generated by C-FAN, we visualize the component-wise quality scores of images from IJB-S dataset. Figure \ref{fig:quality-scores} shows the images of three subjects with the corresponding 512-dimensional quality vectors. We can observe that images with high visual quality have higher quality scores than those with low visual quality. The images with motion blur, occlusion, and deformity tend to be assigned with negative quality values. We also observe that each component of a deep feature vector presents different response to the noise in images.

Figure \ref{fig:failure-examples} shows some failure examples of face identification on IJB-S dataset. For the ``Surveillance-to-Surveillance'' protocol, both probe templates and gallery templates are frames from surveillance videos, making it very difficult to discriminate identities from the images with such low quality and significant variations. However, we observe that the proposed C-FAN tends to retrieve templates with good visual quality. The ground-truth images are more blurry than the retrieved ones in the top two rows of Figure \ref{fig:failure-examples}. Even though probe frames have low visual quality in the ``Surveillance-to-booking'' protocol, we can still observe some facial patterns shared by the probe and the retrieved templates, e.g., the same hair style, the blurry cap and the hair bang. 

\subsection{Quantitative Analysis on IJB-S}

\begin{table*}[!h]
    \centering
    \footnotesize
    \caption{Performance comparisons on IJB-S dataset.}
    \label{table: ijbs}
		\centering
        \scalebox{1.0}{
		\begin{tabularx}{0.9\linewidth}{X c c c c c c c c}
		\toprule
		\multirow{2}{*}{Test Name} & \multirow{2}{*}{Method} && \multicolumn{3}{c}{Closed-set (\%)} && \multicolumn{2}{c}{Open-set (\%)} \\
		\cline{4-6} \cline{8-9}
		& && Rank-1 & Rank-5 & Rank-10 && 1 \% FPIR & 10 \% FPIR \\
		\midrule
        \multirow{3}{*}{Surveillance-to-still}          & Average && $47.93$ & $58.86$ & $63.20$ && $15.22$ & $21.77$ \\
	                                                    & Instance && $49.57$ & $59.58$ & $64.07$ && $\mathbf{16.48}$ & $23.14$ \\
	                                                    & C-FAN && $\mathbf{50.82}$ & $\mathbf{61.16}$ & $\mathbf{64.95}$ && $16.44$ & $\mathbf{24.19}$ \\
		\midrule
		\multirow{3}{*}{Surveillance-to-booking}        & Average&& $49.31$ & $59.92$ & $64.44$ && $26.27$ & $27.05$ \\
		                                                & Instance && $51.14$ & $61.43$ & $65.37$ && $26.64$ & $28.16$ \\
		                                                & C-FAN && $\mathbf{53.04}$ & $\mathbf{62.67}$ & $\mathbf{66.35}$ && $\mathbf{27.40}$ & $\mathbf{29.70}$ \\
		\midrule
		Multi-view & Average                        && $92.57$ & $97.52$ & $99.01$ && $58.42$ & $77.72$ \\
		\multirow{2}{*}{Surveillance-to-booking}        & Instance && $94.55$ & $99.01$ & $99.50$ && $61.39$ & $83.17$ \\
	                                                & C-FAN && $\mathbf{96.04}$ & $\mathbf{99.50}$ & $\mathbf{99.50}$ && $\mathbf{70.79}$ & $\mathbf{85.15}$ \\
		\midrule
		\multirow{3}{*}{Surveillance-to-Surveillance}   & Average && $9.38$ & $17.44$ & $\mathbf{22.14}$ && $0.03$ & $0.54$ \\
		                                                & Instance && $8.90$ & $16.61$ & $21.38$ && $0.06$ & $0.54$ \\
		                                                & C-FAN && $\mathbf{10.05}$ & $\mathbf{17.55}$ & $21.06$ && $\mathbf{0.11}$ & $\mathbf{0.68}$ \\
		\midrule
		UAV & Average                               && $1.27$ & $8.86$ & $13.92$ && $0.00$ & $0.00$ \\
		\multirow{2}{*}{Surveillance-to-booking}        & Instance && $5.06$ & $11.39$ & $15.19$ && $0.00$ & $0.00$ \\
		                                                & C-FAN && $\mathbf{7.59}$ & $\mathbf{12.66}$ & $\mathbf{20.25}$ && $0.00$ & $0.00$ \\
		\bottomrule
		\end{tabularx}}
\end{table*}

Table \ref{table: ijbs} reports identification results on IJB-S dataset. We compare C-FAN with average pooling and instance-level aggregation. Both instance-level aggregation and C-FAN outperforms average pooling on almost all protocols of IJB-S, showing that the features aggregated by quality weights are more discriminative than by simply computing the mean representations. Furthermore, compared with instance-level aggregation, C-FAN achieves higher identification rates on most of the protocols, showing the advantage of the aggregating different features components separately. With the expectation of the Multi-view protocol, the performance of our model on IJB-S dataset is significantly lower than on other datasets (See Section~\ref{sec:ytf_ijba}), showing that this dataset is indeed much more challenging. Notice that the open-set TPIR of ``UAV Surveillance-to-booking'' are all $0\%$. This is because $29$ out of $79$ probe templates are empty (no faces detected) and thus it is impossible to control the FPIR under $36.71\%$.
% In our experiments, both C-FAN and Instance are trained on MS-Celeb-1M, where images are very different from video frames of IJB-S. Thus, C-FAN shows less generalizability than Instance.

\begin{table}[t]
    \centering
    \footnotesize
    \caption{Verification performance on YouTube Face dataset, compared with baseline methods and other state-of-the-art methods.}
    \label{table: ytf}
    \scalebox{1.0}{
    \begin{tabularx}{\linewidth}{X c X c}
        \toprule
        Method & Accuracy (\%) & Method & Accuracy (\%)\\
        \midrule  
        EigenPEP~\cite{li2014eigen} & $84.8 \pm 1.4$ & DeepFace \cite{taigman2014deepface} & $91.4 \pm 1.1$ \\
        DeepID2+~\cite{sun2015deeply} & $93.2 \pm 0.2$ & Wen~\etal~\cite{wen2016discriminative} & $94.9$ \\
        FaceNet~\cite{Schroff_2015_CVPR} & $95.52 \pm 0.06$ & DAN~\cite{rao2017learning} & $94.28 \pm 0.69$ \\
        NAN~\cite{yang2017neural} & $95.72 \pm 0.64$ & QAN~\cite{liu2017quality} & $96.17 \pm 0.09$ \\
        \midrule
        \textit{Average} & $96.36 \pm 1.01$ & \textit{Instance} & $96.42 \pm 0.95$ \\
        && \textit{C-FAN} & $\mathbf{96.50} \pm \mathbf{0.90}$ \\
        \bottomrule
    \end{tabularx}}
\end{table}

\subsection{Performance Comparison on YTF and IJB-A}
\label{sec:ytf_ijba}
% \begin{table}[t]
%     \centering
%     \footnotesize
%     \caption{Verification performance on YouTube Face dataset, compared with baseline methods and other state-of-the-arts.}
%     \label{table: ytf}
%     \scalebox{1.0}{
%     \begin{tabularx}{\linewidth}{X c c}
%         \toprule
%         Method && Accuracy (\%) \\
%         \midrule  
%         EigenPEP \cite{li2014eigen} && $84.8 \pm 1.4$ \\
%         DeepFace-single \cite{taigman2014deepface} && $91.4 \pm 1.1$ \\
%         DeepID2+ \cite{sun2015deeply} && $93.2 \pm 0.2$ \\
%         Wen~\etal~ \cite{wen2016discriminative} && $94.9$ \\
%         FaceNet \cite{Schroff_2015_CVPR} && $95.52 \pm 0.06$ \\
%         DAN \cite{rao2017learning} && $94.28 \pm 0.69$ \\
%         NAN \cite{yang2017neural} && $95.72 \pm 0.64$ \\
%         QAN \cite{liu2017quality} && $96.17 \pm 0.09$ \\
%         \midrule
%         Average && $96.34 \pm 0.96$ \\
%         Instance && $96.38 \pm 0.98$ \\
%         C-FAN && $\mathbf{96.40} \pm \mathbf{1.00}$ \\
%         \bottomrule
%     \end{tabularx}}
% \end{table}

\begin{table*}
    \centering
    \caption{Comparisons of verification and identification on IJB-A dataset.}
    \label{table: ijba}
    \footnotesize
    \begin{tabularx}{\linewidth}{X c c c c c c c c}
        \toprule
        \multirow{2}{*}{Method} & \multicolumn{2}{c}{Verification TAR (\%)} && \multicolumn{2}{c}{Close-set Identification (\%)} && \multicolumn{2}{c}{Open-set Identification (\%)} \\
        \cline{2-3} \cline{5-6} \cline{8-9}
        & 0.1\% FAR & 1\% FAR && Rank-1 & Rank-5 && 1\% FPIR & 10\% FPIR \\
        \midrule
        LSFS \cite{wang2017face} & $51.4 \pm 6.0$ & $73.3 \pm 3.4$ && $82.0 \pm 2.4$ & $92.9 \pm 1.3$ && $38.3 \pm 6.3$ & $61.3\pm 3.2$  \\
        Pose-aware Models \cite{masi2016pose} & $65.2 \pm 3.7$ & $82.6 \pm 1.8$ && $84.0 \pm 1.2$ & $92.5 \pm 0.8$ && - & - \\
        Masi~\etal~ \cite{masi2016we} & $72.5$ & $88.6$ && $90.6$ & $96.2$ && - & - \\
        Triplet Embedding \cite{sankaranarayanan2016triplet} & $81.3 \pm 2$ & $90.0 \pm 1.0$ && $93.2 \pm 1.0$ & - && $75.3 \pm 3.0$ & $86.3 \pm 1.4$ \\
        Shi~\etal~ \cite{shi2018improving} & $60.2 \pm 6.9$ & $82.3 \pm 2.2$ && $89.8 \pm 0.92$ & $96.0 \pm 0.6$ && $58.9 \pm 1.6 $ & $64.7 \pm 1.4$ \\
        QAN \cite{liu2017quality} & $89.3 \pm 3.9$ & $94.2 \pm 1.5$ && - & - && - & - \\
        NAN \cite{yang2017neural} & $88.1 \pm 1.1$ & $94.1 \pm 0.8$ && $95.8 \pm 0.5$ & $98.0 \pm 0.5$ && $81.7 \pm 4.1$ & $91.7 \pm 0.9$ \\
        Pooling Faces \cite{hassner2016pooling} & $63.1$ & $81.9$ && $84.6$ & $93.3$ && - & - \\
        \midrule
        \textit{Average} & $91.14 \pm 1.13$ & $\mathbf{93.99} \pm \mathbf{0.79}$ && $94.44 \pm 0.88$ & $96.22 \pm 0.53$ && $86.23 \pm 4.30$ & $92.36 \pm 1.00$ \\
        \textit{Instance} & $91.52 \pm 0.73$ & $93.78 \pm 0.77$ && $94.54 \pm 0.87$ & $96.24 \pm 0.57$ && $86.58 \pm 4.76$ & $92.45 \pm 0.99$ \\
        \textit{C-FAN} & $\mathbf{91.59} \pm \mathbf{0.99}$ & $93.97 \pm 0.78$ && $\mathbf{94.57} \pm \mathbf{0.82}$ & $\mathbf{96.27} \pm \mathbf{0.53}$ && $\mathbf{86.88} \pm \mathbf{4.70}$ & $\mathbf{92.85} \pm \mathbf{1.02}$ \\
        \bottomrule
    \end{tabularx}
\end{table*}
%B-CNN \cite{chowdhury2016one} & - & - && $58.8 \pm 2.0$ & $79.6 \pm 1.7$ && $14.3 \pm 2.7$ & $34.1 \pm 3.2$ \\
% Deep Multi-pose \cite{abdalmageed2016face} & - & $87.6$ && $84.6$ & $92.7$ && $52.0$ & $75.0$ \\
Table \ref{table: ytf} and Table \ref{table: ijba} report the results of C-FAN, the corresponding baseline methods, and other state-of-the-art on YTF and IJB-A, respectively. C-FAN achieves the best performance on YTF dataset. In particular, C-FAN outperforms the state-of-the-art algorithm NAN \cite{yang2017neural} by $0.78\%$ and QAN by $0.33\%$. On IJB-A dataset, C-FAN achieves the best performance on 1:1 verification and open-set identification tasks, compared to previous methods. It can be seen that C-FAN outperforms NAN by $3.49\%$ at $0.1\%$ FAR, and by $5.18\%$ at $1\%$ FPIR, respectively. Notice that the gaps between C-FAN and two baseline pooling methods are relatively small. This is because the images in YTF and IJB-A datasets are not typical video frames from real-world surveillance scenarios and environments. Most images in these datasets contain rich information for face recognition. Thus, a smaller improvement is expected for quality-aware feature aggregation.

\section{Conclusion}

In this paper, we propose a component-wise feature aggregation network (C-FAN) for video face recognition. It adaptively predicts quality values for each component of a deep feature vector and aggregates the most discriminative components to generate a single feature vector for a set of face images. We empirically demonstrate that the quality scores predicted by C-FAN fit the visual quality of images and are also beneficial in template representation by retaining discriminative components of feature vectors. Our future work will explore an aggregation network that combines different levels of fusion.

{\footnotesize
\bibliographystyle{ieee}
\bibliography{egbib}
}

\end{document}